\RequirePackage{fix-cm}
\documentclass[twocolumn]{svjour3}          
\smartqed  
\usepackage{graphicx}
\usepackage{subfigure}
\usepackage{array}
\usepackage{amsmath}
\usepackage{epsf,epstopdf}
\usepackage{latexsym}
\begin{document}

\title{Automatic 3D liver location and segmentation via convolutional neural networks and graph cut}


\author{Fang Lu         \and
        Fa Wu  \and
        Peijun Hu  \and
        Zhiyi Peng  \and
        Dexing Kong$^\star$ 
}


\institute{Fang Lu \and Fa Wu \and Peijun Hu \and Dexing Kong$^\star$ \at
              School of Mathematical Sciences\\
              Zhejiang University, Hangzhou {\rm310027}, China; \\
              $^\star$\email{dkong@zju.edu.cn.com}           
           \and
           Zhiyi Peng \at
            Department of Radiology\\
            First Affiliated Hospital of Zhejiang University,\\
            Hangzhou {\rm310003}, China
}

\date{}

\maketitle

\begin{abstract}
\label{abstract}
~~~~~~~~\\
\emph{Purpose}~~Segmentation of the liver from abdominal computed tomography (CT) image is an essential step in some computer assisted clinical interventions, such as surgery planning for living donor liver transplant (LDLT), radiotherapy and volume measurement. In this work, we develop a deep learning algorithm with graph cut refinement to automatically segment liver in CT scans.\\
\emph{Methods}~~The proposed method consists of two main steps: (i) simultaneously liver detection and probabilistic segmentation using 3D convolutional neural networks (CNNs); (ii) accuracy refinement of initial segmentation with graph cut and the previously learned probability map.\\
\emph{Results}~~The proposed approach was validated on forty CT volumes taken from two public databases MICCAI-Sliver07 and 3Dircadb. For the MICCAI-Sliver07 test set, the calculated mean ratios of volumetric overlap error (VOE), relative volume difference (RVD), average symmetric surface distance (ASD), root mean square symmetric surface distance (RMSD) and maximum symmetric surface distance (MSD) are 5.9\%, 2.7\%, 0.91\%, 1.88 mm, and 18.94 mm, respectively. In the case of 20 3Dircadb data, the calculated mean ratios of VOE, RVD, ASD, RMSD and MSD are 9.36\%, 0.97\%, 1.89\%, 4.15 mm and 33.14 mm, respectively. \\
\emph{Conclusion}~~The proposed method is fully automatic without any user interaction. Quantitative results reveal that the proposed approach is efficient and accurate for hepatic volume estimation in a clinical setup. The high correlation between the automatic
and manual references shows that the proposed method can be good enough to replace the time-consuming and non-reproducible manual segmentation method.
\end{abstract}

\section*{Introduction}
\label{sec:intro}
Liver diseases pose a serious threat to the health and lives of human beings. Liver cancer has been reported as the second most frequent cause of cancer death in men and the sixth leading cause of cancer death in women. Indeed, about 750,000 people were diagnosed with liver cancer and nearly 696,000 people died from this disease worldwide in 2008~\cite{Jemal2011Global}. Contrast enhanced computed tomography (CT) is now routinely being used
for the diagnosis of liver disease and surgery planning. Liver segmentation from CT is an essential step for some computer assisted clinical interventions, such as surgery planning for living donor liver transplant (LDLT), radiotherapy and volume measurement. Currently, manual delineation on each slice by experts is still the standard clinical practice for the liver delineation. However, manual segmentation is subjective, poorly reproducible, and time consuming. Therefore, it is necessary to develop automatic segmentation method to accelerate and facilitate diagnosis, therapy planning and monitoring.

To date, several methods have been proposed for liver segmentation from CT scans and reviewed in \cite{survey2009comparison}. To summarize, those approaches can be generally classified as: interactive method \cite{Beichel2012Liver}, semi-automatic method \cite{Peng2014A,Pengconvex,Freiman2008An} and automatic method \cite{Massoptier2007Fully,Tomoshige2014A,Wang2015Shape}. Interactive method and semi-automatic method usually need several user guidance or massive interactive operations, which will decrease the efficiency of the physician and undesirable in the practical clinical usage. Thus, fully automatic liver segmentation methods have received extensive attention.
\begin{figure}[!ht]
\label{fig:1}
\centering\includegraphics[width=0.45\textwidth]{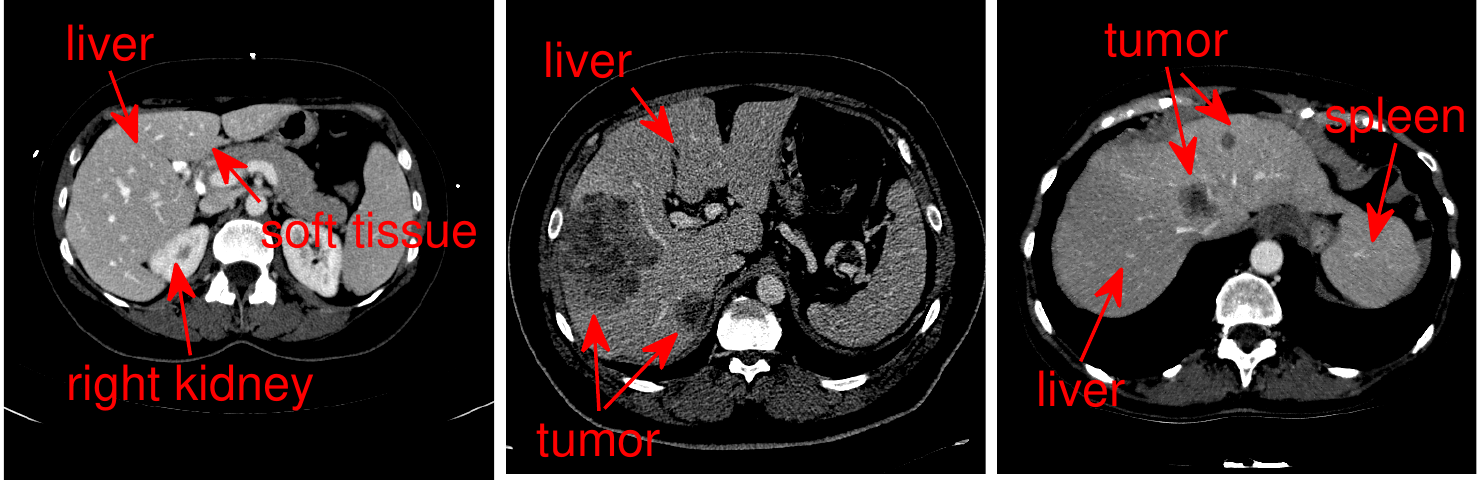}
\caption{Illustration of the challenges in automatic liver segmentation. The liver shares the similar intensity distributions
with its surrounding organs (e.g., the right kidney and the spleen). The shape and appearance of the liver vary largely across
subjects}
\end{figure}
\par
Current automatic liver segmentation can be broadly grouped into two groups: anti-learning based methods and learning based methods. The former mainly includes thresholding \cite{th1}, region growing \cite{reg1}, level set based methods \cite{Lee2007Efficient,Pan2006}, graph cut based methods \cite{Afifi2012Liver,Beichel2012Liver,Chen20113D} and so on. Seo et al. \cite{th1} used several histogram processes, including histogram transformation, multi-modal threshold and histogram tail threshold to automatically segment the liver. Rusk$\acute{o}$ et al. \cite{reg1} incorporated the local neighborhood of the voxel to propose neighbor-hood-connected region-growing for automatic 3D liver segmentation. However, with the low contrast, weak edges and the high noise in CT images, it employed several pre-processing and post-processing steps to decrease under- or over-segmentation.  With the abilities of capturing objects with complex shape and controlling shape regularity, level set methods are attractive in liver segmentation\cite{Pan2006,Al-ShaikhliYR15}. For instance, Al-Shaikhli et al. \cite{Al-ShaikhliYR15} developed a level set method using sparse representation of global and local image information for automatic 3D liver segmentation. Graph cut based methods, the extension of the classic graph cut proposed by Boykov et al. \cite{Boykov2001Interactive,Boykov2006Graph}, are popular in liver segmentation \cite{Beichel2012Liver,Afifi2012Liver,Peng20153D}. Afifi et al. \cite{Afifi2012Liver} proposed a graph cut algorithm based on the iteratively estimated shape and intensity constrains in a slice by slice manner to segment the liver. Massoptier et al. \cite{Massoptier2007Fully} applied a graph cut method initialized by an adaptive threshold to automatically segment the liver on CT and MR images. 
Linguraru et al. \cite{Linguraru2011Liver} integrated a generic affine invariant shape parameterization method into geodesic active contour to detect the liver, followed by liver tumor segmentation using graph cut. Li et al. \cite{Li2015Automatic} proposed a deformable graph cut, which effectively integrated the shape constrains into region cost and boundary cost of the graph cut in a narrow band, to accurately detect the liver surface.
\par Active shape models (ASMs) \cite {ssm1} based methods and atlas based methods \cite{Park2003Construction} are classical learning based methods. ASMs first construct a prior shape of the liver by statistical shape models (SSMs) and then match it to the target image. Kainm$\ddot{u}$ller et al. \cite{ssm3} integrated statistical deformable model to a constrained free-form segmentation method. Heimann et al. \cite{Heimann2007A} presented a fully automated method based on a SSM and a deformable mesh to tackle the liver segmentation task. Wimmer et al. \cite{ssm4} proposed a probabilistic active shape model, which combined boundary, region, and shape information in a single level set equation. Erdt et al. \cite{Erdt2010Fast} proposed a multi-tiered statistical shape model for the liver that combines learned local shape constraints with observed shape deviation during adaptation. Recently, Wang et al. \cite{ssm6} employed the sparse shape composition model to construct a robust shape prior for the liver to help to achieve the accurate segmentation of the liver, portal veins, hepatic veins, and tumors simultaneously. Although the ASMs aforementioned perform well on liver segmentation, they require a complicated and time consuming model construction process. Probabilistic atlas based methods first form the atlas, and then seeks the correspondence between the liver atlas and this structure in the target image by a registration algorithm \cite{Park2003Construction}. However, the precise registration of abdominal CT images is difficult and time consuming. Additionally, atlas selection and label fusion used atlas based method are not easy. Thus, the clinical utility of these methods is limited.

Nevertheless, each of the existing techniques in the literature has limitations, when used on challenging cases. The main challenges may be summarized as follows. First, the liver shares the similar intensity distributions with its surrounding organs (e.g., the heart, the right kidney and the spleen). This makes it more challenging especially for automatic liver detection. Second, the shape and appearance of the liver vary largely across subjects. Finally, the presence of tumors or other abnormalities may result in serious intensity inhomogeneity. Figure 1 illustrates typical challenges as described above.
\begin{figure}[!htbp]
\label{fig:2}
\centering\includegraphics[width=0.55\textwidth]{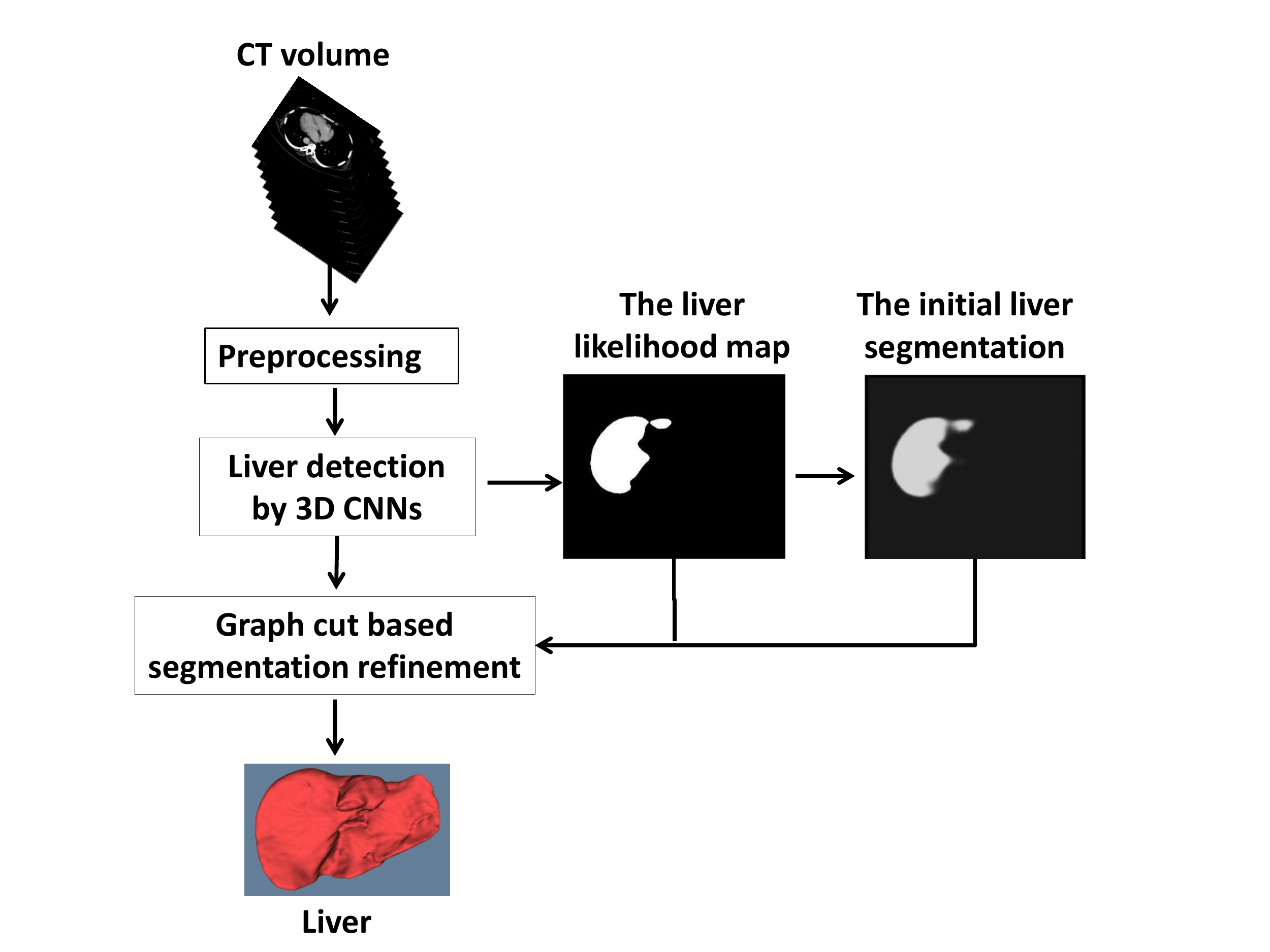}
\caption{Flowchart of the proposed liver segmentation method}
\end{figure}
Recently, deep learning models, which can learn a hierarchy of features by building high level features from low level ones, have received a lot of attention. The CNNs, a classical type of deep learning models, can capture complicated nonlinear mappings between inputs and outputs \cite{Lecun1998Gradient,Krizhevsky2012ImageNet}, which is highly desirable for target detection. Accordingly, superior performance with CNNs has been obtained on many computer vision problems, including visual object recognition \cite{Szegedy2013Deep} and image segmentation \cite{Dan2012Deep,Zhang2015Deep,CERNAZANU2013Segmentation}. For instance, Prasoon et al. \cite{Adhish2013Deep} integrated three 2D CNNs for knee cartilage segmentation in MR images. Zhang et al. \cite{Zhang2015Deep} applied 2D CNNs for multi-modality isointense infant brain image segmentation. Cernazanu et al. \cite{CERNAZANU2013Segmentation} used CNNs in X-ray images to detect bone structure accurately. However, 3D CNNs has not been introduced into the task of liver segmentation from CT scans yet.
\par In this work, we develop a fully automatic liver segmentation framework by utilizing a combined deep learning and graph cut approach. Specifically, it starts by learning the liver likelihood map to automatically identify the liver surface by the generative CNNs model. Then the learned probability map for the liver is incorporated into a graph cut model to refine the initial segmentation. We evaluate the proposed method on 40 contrast enhanced CT volumes from two public databases. In terms of novelty and contributions, our work is one of the early attempts of employing 3D CNNs for liver segmentation. The proposed method can simultaneously learn low level features and high level features. Moreover, the proposed approach is fully automatic without any user interaction. Thus it can increase the efficiency of the physician.
\section*{Datasets}
\label{sec:data}
\subsection*{Training dataset}
78 contrast-enhanced CT volumes with ground truth are collected in the transversal direction. Among them, 10 are from the MICCAI-Sliver07 training dataset\footnote{In detail, they are the liver002, liver004, liver006, liver008, liver010, liver012, liver014, liver016, liver018 and liver020.}, while other 68 volumes from our partner site with ground truth given by experienced experts. There are 26 abnormal livers and 52 normal livers. The pixel spacing varies between 0.55 mm and 0.81 mm, whereas inter-slice distance varies from 0.7 mm to 3 mm and slice number 64 to 346.
\subsection*{Test dataset}
The test datasets consist of 40 contrast-enhanced CT volumes with 512$\times$512 in-plane resolution. Among them, 10 are from the MICCAI-Sliver07 training set\footnote{In detail, they are the liver001, liver003, liver005, liver007, liver009, liver011, liver013, liver015, liver017 and liver019.}, 10 are from the MICCAI-Sliver07 test dataset, and 20 are from the public database 3Dircabd. The pixel size varies from 0.54 mm to 0.86 mm, slice thickness from 0.7 mm to 5 mm, and slice number 64 to 502.
\begin{table*}
\label{table:CNN}
\caption{Detailed architecture of 3D CNNs used in this work. \emph{Conv} and \emph{Norm} denote convolutional layers and normalization layers, respectively}
\centering
\begin{tabular}{lcccc}
\hline
 Layer   & Input     & filter     & padding    &  Output\\
  \hline
  $Conv$ $\longrightarrow$ $Norm$   & 249$\times$249$\times$279$\times$1     & 7$\times$7$\times$9$\times$96     & 3$\times$3$\times$0     &  125$\times$125$\times$136$\times$96  \\

  $Pooling$& 125$\times$125$\times$136$\times$96     & 3$\times$3$\times$2     & 1$\times$1$\times$0     &  63$\times$63$\times$68$\times$96  \\

  $Conv$& 63$\times$63$\times$68$\times$96     & 5$\times$5$\times$5$\times$256     & 2$\times$2$\times$0     &  63$\times$63$\times$64$\times$256  \\

  $Pooling$& 63$\times$63$\times$64$\times$256     & 3$\times$3$\times$2     & 0$\times$0$\times$0     &  31$\times$31$\times$32$\times$256  \\

  $Conv$& 31$\times$31$\times$32$\times$256      & 3$\times$3$\times$3$\times$512     & 1$\times$1$\times$1     & 31$\times$31$\times$32$\times$512  \\

  $Conv$& 31$\times$31$\times$32$\times$512    & 3$\times$3$\times$3$\times$512     & 1$\times$1$\times$1    & 31$\times$31$\times$32$\times$512   \\

  $Conv$& 31$\times$31$\times$32$\times$512    & 3$\times$3$\times$3$\times$512     & 1$\times$1$\times$1    &  31$\times$31$\times$32$\times$512   \\

  $Conv$& 31$\times$31$\times$32$\times$512    & 3$\times$3$\times$3$\times$512     & 1$\times$1$\times$1    & 31$\times$31$\times$32$\times$512   \\

  $Conv$& 31$\times$31$\times$32$\times$512    & 3$\times$3$\times$3$\times$512     & 1$\times$1$\times$1    &  31$\times$31$\times$32$\times$512   \\

 $Double$ $size$& 31$\times$31$\times$32$\times$512      & -    & -    &  62$\times$62$\times$64$\times$64  \\

 $Conv$&  62$\times$62$\times$64$\times$64      & 3$\times$3$\times$3$\times$512     & 1$\times$1$\times$1     &  62$\times$62$\times$64$\times$512  \\

 $Double$ $size$& 62$\times$62$\times$64$\times$512     & -    & -    &  124$\times$124$\times$128$\times$64  \\

 $Conv$& 124$\times$124$\times$128$\times$64     & 3$\times$3$\times$3$\times$128     & 1$\times$1$\times$1     &  124$\times$124$\times$128$\times$128  \\

 $Double$ $size$& 124$\times$124$\times$128$\times$128     & -    & -     &  248$\times$248$\times$256$\times$16  \\

  $Conv$& 248$\times$248$\times$256$\times$16     & 3$\times$3$\times$3$\times$16     & 1$\times$1$\times$1     &  248$\times$248$\times$256$\times$16  \\

  $Conv$$\longrightarrow$$Logistic$& 248$\times$248$\times$256$\times$16     & 3$\times$3$\times$3$\times$1     & 1$\times$1$\times$1     & 248$\times$248$\times$256$\times$1  \\
  \hline
\end{tabular}
\end{table*}
\section*{Method}
\label{sec:method}
A flowchart of the proposed method is depicted in Fig. 2. The proposed method consists of two main parts: 3D deep CNNs based liver detection and 3D graph cut based segmentation refinement.
\subsection*{3D deep CNNs based liver detection and segmentation}
\subsubsection*{Introduction of CNNs}
\label{ssec:CNN intro}
We just briefly review the method of CNNs in this section. More information about this network can be found in the literature~\cite{Krizhevsky2012ImageNet,Zeiler2014Visualizing}. CNNs is a variation of multilayer perceptron. The convolutional layers and subsampling layers are core blocks of CNNs. Several convolutional layers can be stacked on top of each other to learn a hierarchy of features. Each convolutional layer is used to extract feature maps of its preceding layer, which is connected by some filters. We denote $C^{(m-1)}$ and $C^{(m)}$ as the input and output for the $m$-th convolutional layer, respectively, and $C_i^{(m)}$ the $i$-th output feature map of the $m$-th layer. The outputs of the $m$-th layer can be computed as,
\begin{equation}
C_j^{(m)}=F_{W,b}(\sum_i{C_i^{(m-1)}*w_{ij}^{(m)}+b_j^m});
\end{equation}
where $*$ denotes the convolution, $w_{ij}^{(m)}$ denotes the kernel linking the $i$-th input map and the $j$-th
output map and $b_j^{(m)}$ is the bias for the $j$-th output map in the $m$-th layer. $F_{W,b}(\cdot)$ is a nonlinear activation function. There are multiple choices for it, such as the sigmoid, hyperbolic tangent and rectified linear functions. In order to reduce the computational complexity and introduce invariance properties, a subsampling layer is often used after a convolutional layer. As for the pooling layer, which is a common subsampling layer, we adopt the average pooling, which uses mean values within 3$\times$3 groups of pixels centered at the pooling unit, with the distance between pooling set to two pixels. The final convolutional layer is usually followed by the softmax classifier. For the binary classification problem, logistic regression is used to normalize the result of the kernel convolutions into a multinomial distribution over the labels. The major advantage of the convolutional networks is the use of shared weights in convolutional layers, which means that the same filter is used for each pixel in the layer; this not only reduces the required memory size but also improves the performance.
\par
Assume the training set is made up of $n$ labeled samples $\{(x^1,y^1), (x^2,y^2),...,(x^n,y^n)\}$, where $y^i=0$ or $1$, $i=1,2,\cdots,n$. Denote $\theta$ be the set of all the parameters including the kernel, bias and softmax parameters of the CNNs. For logistic regression, we need to minimize the following cost function with respect to $\theta$,
\begin{equation}
E(\theta)=-\frac{1}{n}[\sum_{i=1}^{i=n}{y^i logF_{\theta}(x^i)+(1-y^i)log(1-F_{\theta}(x^i))}].
\end{equation}
We use weight decay, which penalizes too large values of the softmax parameters, to regularize
the classification. The cost function is minimized by gradient-based optimization \cite{Lecun1998Gradient} and the partial derivatives are computed using backpropagation \cite{Krizhevsky2012ImageNet}.
\subsubsection*{Architecture of the proposed 3D CNNs}
\label{ssec:architecture}
As described above, the capacity of CNNs varies, depending on the number of layers. The more layers the network has, the higher level features it will capture. Focusing on the feasibility of the CNNs in liver segmentation, we only provide one architecture of 3D CNNs as detailed in Table~1. The architecture of proposed 3D CNNs contains one input feature map corresponding to CT image block of 249$\times$249$\times$279. It then stacks eleven convolutional layers by some filters, and each layer is followed by the rectified linear unit \cite{Krizhevsky2012ImageNet} to expedite the training of CNNs. This network also uses pooling and softmax layers.
\par
The first convolutional layer contains 96 feature maps. Each of the maps is linked to the input feature maps through filters of size 7$\times$7$\times$9. Then a stride size of two voxels is used to generate feature maps of size 125$\times$125$\times$136. A local response normalization
scheme is applied after the first convolution layer. Following the normalization layer, the mean pooling layer has 96 feature maps of size 63$\times$63$\times$68. The second convolution layer takes the output of the pooling layer as input containing 256 feature maps. Each of the feature maps is linked to all of the feature maps in the previous layer by filters of size of 5$\times$5$\times$5. A stride size of one voxel and the mean pooling layer are used to generate
256 feature maps. The following 5 convolutional layers have 512 feature maps of size 31$\times$31$\times$32. They are connected to all feature maps in the previous layer by
 3$\times$3$\times$3 filters. In addition to convolutional layers, rearranging layers are used before the following three convolution layers, converting 8 channels into 2$\times$2$\times$2, i.e., doubling dimensions and 1/8 channel. The rearranging skill can obtain unambiguous boundaries while upsampling can not. Thus convolution layer after rearranged layer can eliminate blocking artifacts. And the last rearranging layer gives 16 feature maps of 248$\times$248$\times$256. The output of the log-regression layer at last ranges from 0 to 1, which can be interpreted as the probability of each voxel $x$ in the output image block 248$\times$248$\times$256 being classified.
\begin{figure}[!h]
\label{fig:3}
\centering\includegraphics[width=0.45\textwidth]{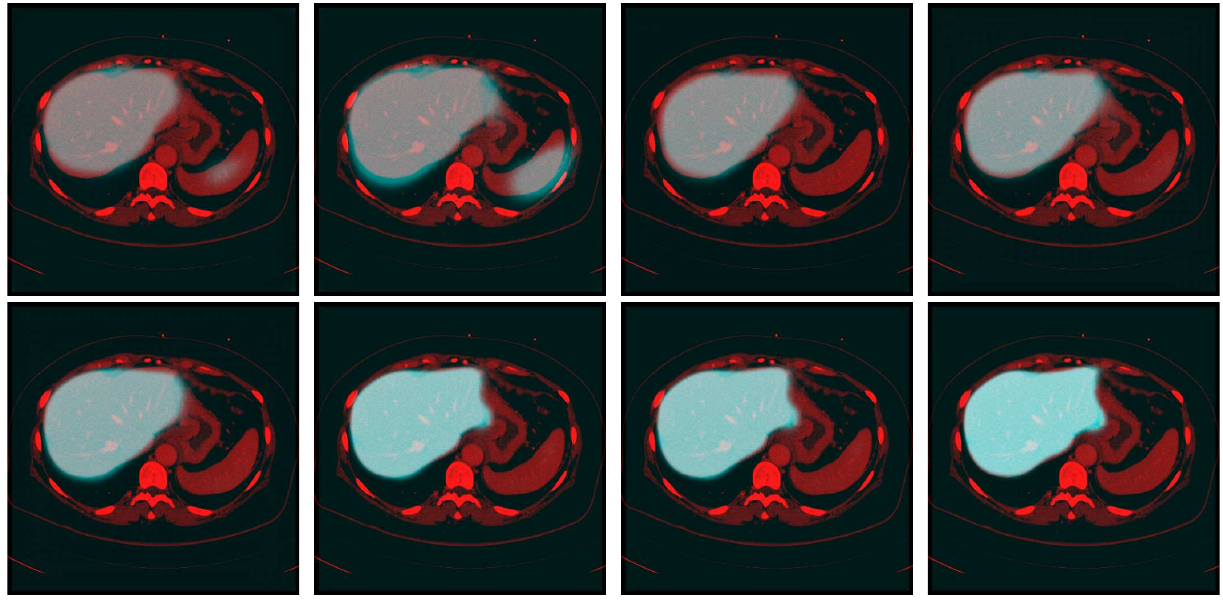}
\caption{Iterative results of the liver likelihood map generated by 3D CNNs for one CT volume of database 3Dircabd. From top left to bottom right, the 3rd, 8th, 13th, 20th, 27th, 34th, 42th, and 53th iterative liver likelihood maps are shown. The brighter the region is, the greater the probability of the liver region is}
\end{figure}
\begin{figure}[!h]
\label{fig:4}
\centering\includegraphics[width=0.45\textwidth]{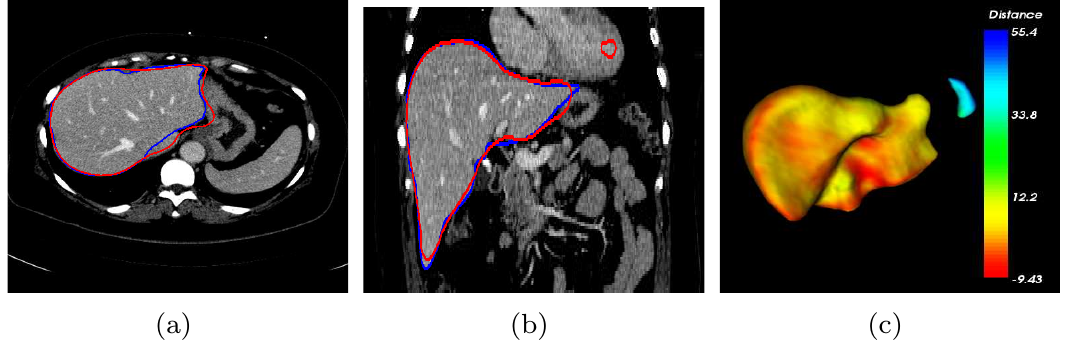}
\caption{Illustrations of the liver region located by 3D CNNs. The located liver region by 3D CNNs is in red with ground truth in blue. \textbf{a} 2D view in axial plane, \textbf{b} 2D view in coronal plane, \textbf{c} the surface distance error (mm) of 3D CNNs with the ground truth }
\end{figure}
\subsection*{Graph cut based segmentation refinement}
\label{ssec:graph cut}
We develop a combined method that uses the CNNs liver likelihood map and graph cut to segment the
liver from the surrounding tissue. The method is initialized by the rough liver region generated by the liver likelihood map.
\par Let us denote $I : x \in \Omega \to R$ a CT volume defined on the domain
$\Omega \subset {R^3}$, $V$ the set of voxels in $\Omega$ and
$N_x$ the standard 6-connected neighborhood of voxel $x$ in 3D grid. Let $l_x \in \{0,1\}$ be the label assigned to voxel $x$,~ where 0 and 1 stand for the background (non-liver region) and the object (liver region), respectively. The aim of the proposed model is to find a label $l = \{l_x,~x\in V\}$ which minimizes the general energy function as follows,
\begin{equation}
\begin{split}
E(l)
&=\lambda E_{D}(l)+ E_{B}(l)\\
&=\lambda\sum_{x\in V}{D_x(l_x)+\sum_{x\in V}\sum_{y\in N_x}{B_{xy}(x,y)\delta(l_x,l_y)}},\end{split}
\end{equation}
where
\begin{equation}
\delta_{xy}(l_x,l_y)=
\left\{ \begin{aligned}
          &1, & if~~l_x\neq l_y,\\
          &0,&otherwise.
          \end{aligned} \right.
\end{equation}
and the coefficient $\lambda$ controls the balance between the data fitting term $E_D(l)$ and the boundary penalty $E_B(l)$. The regional cost term $D_x(l_x)$ describes the degree of similarity between voxel $x$ and the foreground or the background, while the boundary cost term $B_{xy}(x,y)$ encodes the discontinuity between the two
neighboring voxels $x$ and $y$. Both of them have been defined in various ways by different researchers~\cite{Boykov2001Interactive,Boykov2006Graph,Afifi2012Liver}. We define the boundary term $B_{xy}$ as,
\begin{equation}
B_{xy}(x,y)=\frac{1}{1+\beta |I(x)-I(y)|^2},
\end{equation}
where $\beta$ is a constant. The special form of the data term we adopt will be detailed in the following part.
\par
As described above, the data penalty term usually reflects the degree of similarity between voxels and the foreground or the background.~From the initial segmented liver region $L_0$ by 3D CNNs,~an~intensity range $[\zeta,\eta]$ of liver can be roughly estimated as in \cite{Peng2014A}.~Then the thresholding map reads as,
\begin{equation}
f(x) = \frac{(I(x) - \zeta )(I(x) - \eta )}{(\eta  - \zeta )^2}.
\end{equation}
We also~introduce a local appearance term represented by the distribution of a group of features as in \cite{Peng2014A}. Three complementary features, the image intensity $I(x)$, the modified local binary pattern $LBP^{\tau}_{P,r}$ and the local variance of intensity $VAR_{P,r}$, are picked to form a joint feature $F_I(x)=(I(x), LBP^{\tau}_{P,r}, VAR_{P,r})$. In detail,
\begin{equation}
LBP_{P,r}^\tau  = \sum\limits_{p = 0}^{P - 1} {H({I_p} - {I_c} - \tau  \cdot sign({I_p} - {I_c}))} {2^p},
\end{equation}

\begin{equation}
VA{R_{P,r}} = \frac{1}{P}\sum\limits_{p = 0}^{P - 1} {{{({I_p} - {I_m})}^2}} ,{I_m} = \frac{1}{P}\sum\limits_{p = 0}^{P - 1} {{I_p}},
\end{equation}
where $I_p (p=0,1,...,P-1)$ correspond to the intensities of $P$ equally spaced voxels on a sphere of radius $r$, forming a spherically symmetric neighbor set and $I_c$ is the intensity of the center voxel. $H(x)$ is the Heaviside function.
Let $H_x^i$ be the cumulative histogram of the $i$th feature at $x$ in a local window $O(x)$, $H_0^i$ be the mean cumulative histogram of the $i$th feature
on $L_0$ with its variance $\sigma_0^i$. Then a local appearance map reads as,
\begin{equation}
\mathcal{P}(x)=\sum\limits_{i=1}^{i=3}{\frac{\mathcal {W}^1(H_x^i,H_0^{i})}{(\sigma_0^i)^2}},
\end{equation}
here $\mathcal {W}^1(\cdot,\cdot)$ is the $\mathcal {L}1$ Wasserstein distance \cite{Ni2007Local}. By combining the probability map $L(x)$, the thresholding map $f(x)$ and the local appearance map~$\mathcal{P}(x)$,~the data term $D_x(l_x)$ is computed as following,
\begin{equation}
D_x(l_x)=max(-\mathcal{R}(x),0)l_x+max(\mathcal{R}(x),0)(1-l_x),
\end{equation}
where
\begin{equation}
\mathcal{R}(x)=\sum_{y\in N_x}{B_{xy}(x,y)[f(x)+L(x)-0.5+\gamma \mathcal{P}(x)]}
\end{equation}
with $\gamma$ a positive trade-off coefficient.
\par To minimize the total energy function defined as (3) by the graph cut
algorithm, the corresponding graph in 3D grid is defined as follows. Let $G(V,e)$ be the
undirected weighted graph with a set of directed edges $e$ connecting neighboring nodes.
There are also two specially designated  special nodes that are
called terminals, the source $S$ and the sink $T$.~Generally, there are two types of edges in the graph: $n$-links and $t$-links. $n$-links stand for edges between neighboring voxels, while $t$-links are used to connect voxels to terminals. Then, the graph $G$ with cut cost equaling the value of $E(l)$ is constructed using the edge weights defined as follows,
\begin{equation}
e_{sx}=\left\{ \begin{aligned}
          &D_x(l_x=0), & if~~\mathcal{R}(x)>0,\\
          &0,&otherwise.
          \end{aligned} \right.
\end{equation}
\begin{equation}
e_{xt}=\left\{ \begin{aligned}
          &D_x(l_x=1), & if~~\mathcal{R}(x)<0,\\
          &0,&otherwise.
          \end{aligned} \right.
\end{equation}
\begin{equation}
e_{xy}=B_{xy}(x,y),
\end{equation}
$e_{sx}$, $e_{xt}$ are the weights of the links to terminal nodes, and $e_{xy}$ is the weight of the link between two adjacent voxels.
\par In fact, the proposed model is inspired by the Region
Appearance Propagation (RAP) model proposed in \cite{Peng2014A}. However, there are three main improvements as follows. First, the RAP model is proposed in the continuous form and optimized by the level set method. With a gradient decent method for optimization, the solution of the level set is often local,~while that of graph cut referred by us is global. Second, the RAP model needs users to draw the initial region inside the liver to form the initial surface and compute some statistical features. The user intervention may reduce their usability due to the consumption of clinician's time and make the final results be user-dependent. In our paper, an automatic initialization
of a large initial region is generated by the preceding deep learning step. Last but most important, the most liver likely region generated by 3D CNNs
is integrated into the image data penalty term $D_x(l_x)$ to overcome the deficiencies of RAP, such as lack of global information and difficulty in capturing complex texture features. Indeed, this study effectively combines the advantages of RAP and 3D CNNs to develop an automatic and accurate liver segmentation approach.
\begin{figure}[!h]
\label{fig:5}
\centering\includegraphics[width=0.45\textwidth]{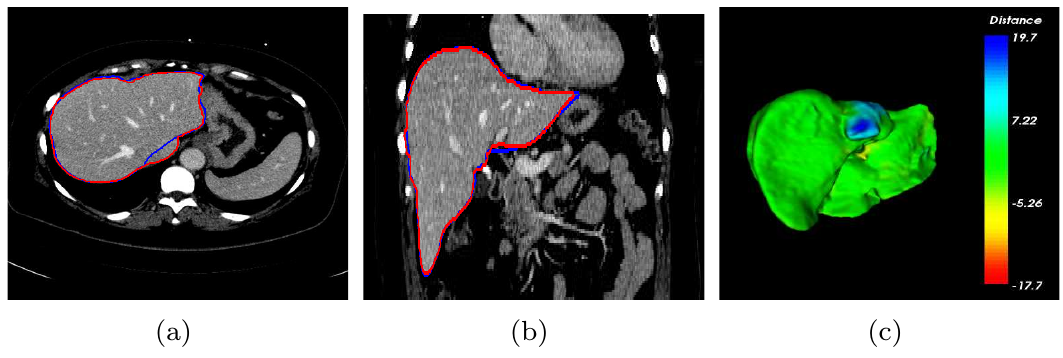}
\caption{Illustrations of the segmentation results by the proposed method. The contour of the segmentation result is in red with ground truth in blue. \textbf{a} 2D view in axial plane, \textbf{b} 2D view in coronal plane, \textbf{c} the surface distance error (mm) of the proposed with the ground truth}
\end{figure}
\begin{figure}[!ht]
\label{fig:6}
\centering\includegraphics[width=0.45\textwidth]{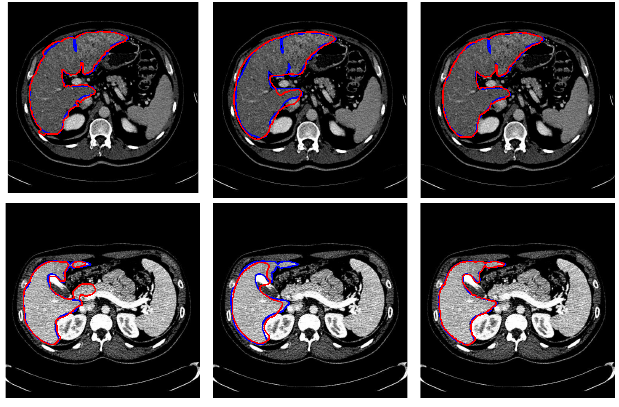}
\caption{Illustrations of the role of the likelihood liver map. From the first column to the last, outcomes of graph cut without the likelihood liver map, convolutional neural networks
 and the proposed integrated model for two typical images are displayed respectively in red. The ground truth is in blue}
\end{figure}

\begin{figure*}[!ht]
\label{fig:7}
\centering\includegraphics[width=0.75\textwidth]{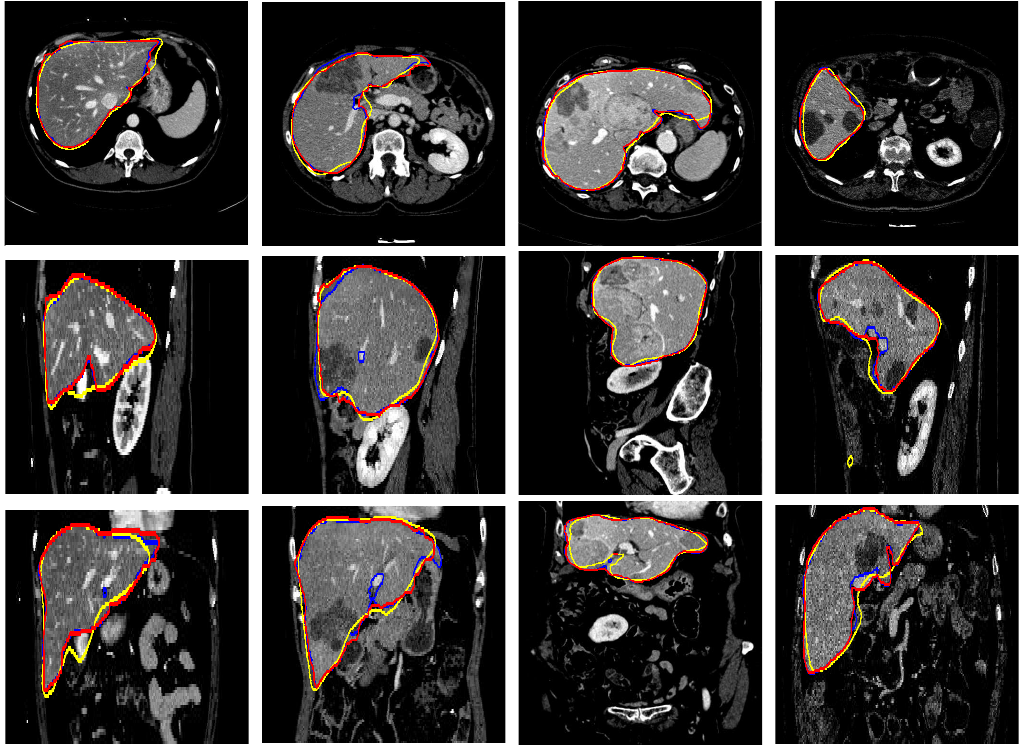}
\caption{2D images of segmentation results of four challenging cases in axial, sagittal, and coronal planes with the ground truth in blue. The initial liver region generated by CNNs is in yellow and the final refined result is in red}
\end{figure*}
\begin{figure*}[!ht]
\label{fig:8}
\centering\includegraphics[width=0.75\textwidth]{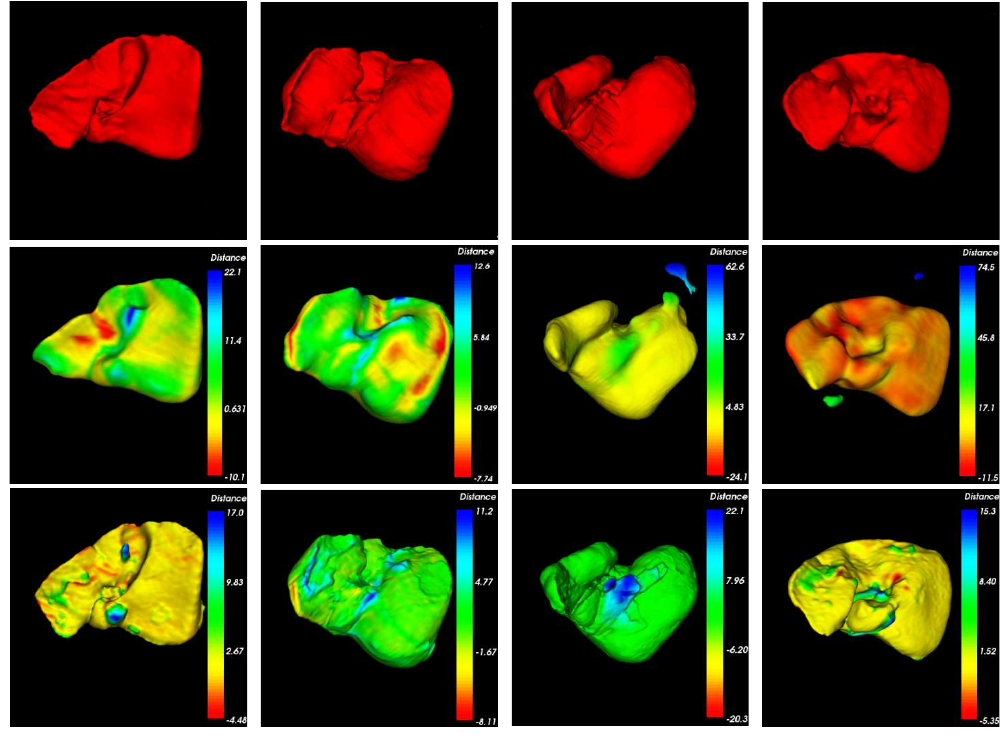}
\caption{3D visual representation of livers segmented by our method
on the same four cases as shown in Fig.~7. The first row shows the ground truth; the second and third rows present the surface distance error (mm) of 3D CNNs and the proposed segmentations with the ground truth}
\end{figure*}
\begin{table*}[!htbp]
\begin{center}
\caption{Evaluation of the proposed method based on MICCAI-Sliver07 test set}
\label{table:test on Sliver07}
\begin{tabular}{c|cc|cc|cc|cc|cc|c}
\hline
Test & VOE &Score &RVD &Score & ASD &Score &RMSD&Score &MSD&Score&Total\\
 case& (\%) & -&(\%) &- & (mm) &- &(mm)&- &(mm)& -&Score\\
\hline
1  &5.29 &76.9 &2.84 &84.9 &0.87 &78.2 &1.68 &76.7  &15.94 &79.0 &79.1 \\
2  &6.95 &72.9 &5.77 &69.3 &1.02 &74.4 &2.18 &69.7  &22.33 &70.6 &71.4 \\ 3  &4.97 &80.6 &0.59 &96.8 &0.92 &76.9 &1.63 &77.4  &13.37 &82.4 &82.8 \\
4  &6.35 &75.2 &2.57 &86.3 &1.09 &72.7 &2.56 &64.5  &26.10 &65.7 &72.9 \\
5  &5.95 &76.8 &0.30 &98.4 &1.04 &73.9 &2.29 &68.1  &25.45 &66.5 &76.7 \\
6  &7.88 &69.2 &4.19 &77.7 &1.18 &70.4 &2.89 &59.8  &27.84 &63.4 &68.1 \\
7  &3.23 &87.4 &0.56 &97.0 &0.43 &89.2 &0.93 &87.1  &13.67 &82.0 &88.6 \\
8  &6.50 &74.6 &5.25 &72.1 &1.08 &73.1 &1.88 &73.9  &14.16 &81.4 &75.0 \\
9  &5.36 &79.1 &3.32 &82.4 &0.60 &85.1 &1.09 &84.8  &15.28 &79.9 &82.2 \\
10  &5.85 &77.1 &1.63 &91.4 &0.83 &79.2 &1.71 &76.2  &15.21 &80.0 &80.8\\
\hline
\textbf{Avg}  &\textbf{5.90} &\textbf{77.0} &\textbf{2.70} &\textbf{85.6} &\textbf{0.91} &\textbf{77.3} &\textbf{1.88} &\textbf{73.8 } &\textbf{18.94} &\textbf{75.1}&\textbf{77.8} \\
\hline
\end{tabular}
\end{center}
\end{table*}
\begin{table*}[!htbp]
\begin{center}
\caption{Comparison with state-of-the-art automatic methods on MICCAI-Sliver07 test set}
\label{table:Comparision on Sliver07}
\begin{tabular}{l|cc|cc|cc|cc|cc|c}
\hline
Method & VOE &Score &RVD &Score & ASD &Score &RMSD&Score &MSD&Score&Total\\
Unit & (\%) & -&(\%) &- & (mm) &- &(mm)&- &(mm)& -&Score\\
\hline
Li et al.~\cite{Li2015Automatic}  &6.24 &- &1.18 &- &1.03 &- &2.11 &-  &18.82 &- &- \\
Shaikhli et al.~\cite{Al-ShaikhliYR15}  &6.44 &74.9 &1.53 &89.7 &0.95 &76.3 &1.58 &78.1  &15.92 &79.1 &79.6 \\
Kainm$\ddot{u}$ller et al. \cite{ssm3} &6.09 &76.2 &-2.86 &84.7 &0.95 &76.3 &1.87 &74.0  &18.69 &75.4 &77.3
\\
Wimmer et al. \cite{ssm4}  &6.47 &74.7 &1.04 &86.4 &1.02 &74.5 &2.00 &72.3  &18.32 &75.9 &76.8  \\
Linguraru et al. \cite{Linguraru2011Liver} &6.37 &75.1 &2.26 &85.0 &1.00 &74.9 &1.92 &73.4  &20.75 &72.7 &76.2 \\
Heimann et al. \cite{Heimann2007A} &7.73 &69.8 &1.66 &87.9 &1.39 &65.2 &3.25 &54.9  &30.07 &60.4 &67.6 \\
Kinda et al. \cite{saddi} &8.91 &65.2 &1.21 &80.0 &1.52 &61.9 &3.47 &51.8  &29.27 &61.5 &64.1 \\
\hline
The proposed  &\textbf{5.90} &77.0 &2.70 &85.6 &\textbf{0.91} &77.3 &1.88 &73.8  &18.94 &75.1 &77.8 \\
\hline
\end{tabular}
\end{center}
\end{table*}
\section*{Segmentation Procedures}
\label{sec:Segmentation Procedures}
The proposed segmentation process contains three stages, i.e., preprocessing, location of the initial liver region, and segmentation refinement. Details of these stages will be described as follows.
\subsection*{Preprocessing}
\label{preprocessing}
Since CNNs are able to learn useful features from scratch, we apply only minimal preprocessing, including three steps. First, to reduce computational complexity, all volumes are resampled 256$\times$256$\times$286 after appending or deleting some slices without liver. Second, the intensity range of all the volumes is normalized to [-128,128] by adjusting the window width and window level. Finally, a 3D anisotropic diffusion filter \cite{Weickert1998Efficient} is used for reducing noise. All the preprocessed steps are applied to both training and test datasets.
\subsection*{Location of the initial liver region}
\label{location of the liver}
Before using the network for locating the liver, it should be trained using the cases in the training set. The CNNs is trained for 53 iterations to generate the liver likelihood map. We observe that after the 13th iteration, the heart and spleen, similar to the liver in terms of intensity or texture, can be differentiated from the liver, as shown in Fig. 3. At around the 40th iteration, the validation result converges. During each iteration, a 249$\times$249$\times$279 block is randomly chosen as the input from a training data, while a 248$\times$248$\times$256 labeled block as the output. We train the parameters of the proposed 3D CNNs by gradient-based optimization. The partial derivatives are computed using backpropagation \cite{Krizhevsky2012ImageNet}. We set the learning rate to 0.1/(248$\times$248$\times$256) at the beginning, and reduce it from 0.1 to 0.005 after the 20th iteration. For other parameters including weight, momentum and decay, we adopt the same as Krizhevsky's \cite{Krizhevsky2012ImageNet}.  Training the network takes approximately 20 hours using 4 pieces of GTX980 GPUs.

After the training, the probability map of liver can be iteratively learned by the trained 3D CNNs. Fig. 3 illustrates the iterative probability map for a test volume. Then, by thresholding, the initial liver shape $L_0$ is easily located, as shown in red in Fig. 4.
\subsection*{Segmentation refinement}
In this step, the liver probability map is used to automatically initialize graph cut and incorporated into the energy function to achieve an accurate result.

From the initial liver shape $L_0$, the intensity range for liver can be roughly estimated as $[\zeta,\eta]=[m-3\sigma,m+3.5\sigma]$, where $m$, and $\sigma$ are the intensity mean and variance over $L_0$, respectively. In the practical usage, parameters used in graph cut are chosen as follows. The balancing weight $\lambda=70$, $\gamma=\sum_{i=1}^{3}{{\sigma_0}^2}/36$, $\beta=0.2$; the local window $O(x)$ is chosen as a cube window of $9\times9\times5$ and the LBP parameters are chosen as $\tau=1.5$, $P=6$, $r=1$. The graph cut segmentation is implemented with C++ on a desktop computer with an Intel Core i5-4460U CPU (3.20 GHz) and a 8 GB of memory. Fig.~5 shows the final segmentation of the case as shown in Fig. 4.
 For a test volume with size of $512\times512\times n$ $(n<286)$,generating the liver likelihood map by 3D CNNs usually consumes about 4s and the graph cut segmentation varies from 20s to 180s.
\section*{Experiments and discussion}
\subsection*{Evaluation metrics}
\label{ssec:metrics}
Five measures of accuracy are calculated as in \cite{survey2009comparison}, i.e., Volumetric Overlap Error (VOE), Relative Volume Difference (RVD), Average Symmetric Surface Distance (ASD), Root Mean Square Symmetric Surface Distance (RMSD) and Maximum Symmetric Surface Distance (MSD). The RVD is given as a signed number to show if the method tend to under- or over-segment.~A perfect scoring result (zero for all the five metrics) is worth 100 per metric, while the manual segmentation by a non-expert of the average quality (6.4\%, 4.7\%, 1 mm, 1.8 mm, and 19 mm) is worth 75 per metric \cite{survey2009comparison}. This segmentation may be regarded as approximately equivalent to the human performance. The final score is the average of the five metric scores.
\par In addition, as a clinical index, liver volumes (LV) are computed for the correlation and Bland-Altman analyses \cite{bland} between the automatic liver segmentation and manual liver segmentation results. The correlation analysis is performed using the least square method to obtain the slope and intercept equation. And the correlation coefficient $R$ is computed. To assess the intra- and inter-observer variability the coefficient of variation (CV), defined as the standard
deviation (SD) of the differences between the automatic and manual results
divided by their mean values is computed.
\subsection*{Results and discussion}
\label{ssec:results and discussion}
To better understand the role of the learned liver likelihood map, Fig. 6 depicts the outputs of the graph cut without the liver likelihood map, 3D CNNs and the proposed method for two typical images in red. The ground truth segmentations drawn by experts are in blue. Obviously, incorporated with the liver likelihood map, the proposed model can achieve a better agreement with the ground truth.

Figure 7 illustrates our segmentation and manual delineations for four challenging cases in coronal, sagittal, and axial planes. The initial liver region generated by 3D CNNs is in yellow, the final refined result is in red and the manual delineation is in blue.
 The first column shows a case with highly inhomogeneous appearances. The last three columns display three representative livers containing tumors. Particularly, some tumors locate on the boundary, which makes it more difficult to automatically delineate the accurate boundary. As can be seen, 3D CNNs can detect the most liver region and the refinement model can obtain a higher agreement with the ground truth. Figure 8 depicts the corresponding 3D visualization results of 3D CNNs and the proposed method for the cases shown in Fig.~7. The 3D visualization of errors is based on the MSD error between the segmentation result and the ground truth. As can been seen, the MSD errors of the 3D CNNs for the four cases (from left to right) are 22.1 mm, 12.6 mm, 62.6 mm and 74.5 mm, respectively, while the MSD errors of the proposed model are 17.0 mm, 11.2 mm, 22.1 mm and 15.3 mm, respectively. Obviously, the proposed approach can obtain lower errors in terms of MSD.
\begin{figure}[!ht]
\label{fig:9}
\centering\includegraphics[width=0.45\textwidth]{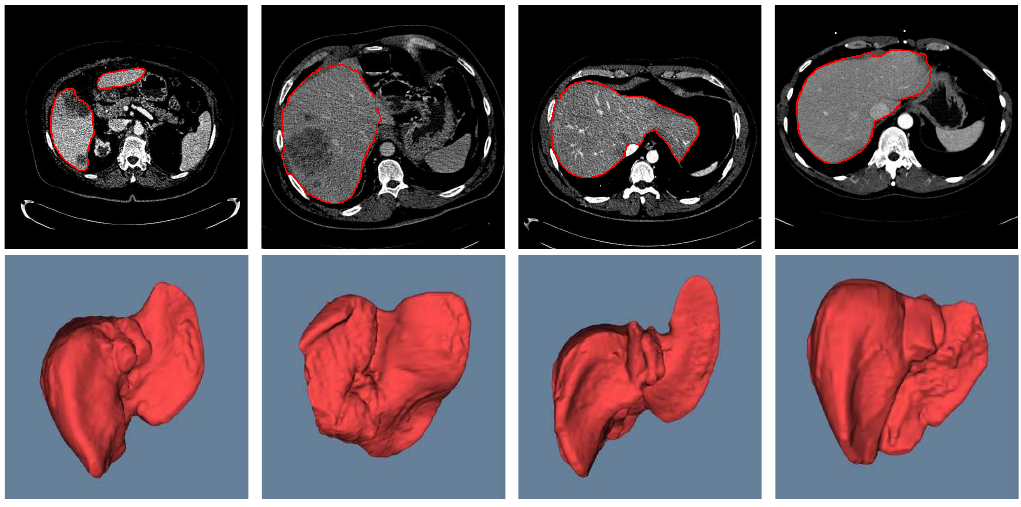}
\caption{Four liver segmentation examples using the MICCAI-SLiver07 test data.
The first row represents the segmentation results of the proposed method in axial plane. The second row shows the 3D visual representations of the final liver segmentation}
\end{figure}

To compare the performance of the proposed framework with state-of-the-art automatic segmentation methods, two tests are conducted on the MICCAI-Sliver07 test set and 3Dircadb database. In the first test, we submit the results on the MICCAI data to the MICCAI-Sliver07 challenge website and the evaluation is obtained by the organizers. Table 2 summarizes the corresponding results in terms of five metrics (VOE, RVD, ASD, RMSD, and MSD). The calculated mean ratios of VOE, RVD, ASD, RMSD, and MSD are 5.9\%, 2.7\%, 0.91\%, 1.88 mm, and 18.94 mm, respectively. Figure 9 presents the results of four typical liver examples. Table~3 lists the comparative results of the proposed approach and the other eight fully automatic methods \cite{Li2015Automatic,Al-ShaikhliYR15,ssm3,ssm4,Linguraru2011Liver,Heimann2007A,saddi} based on MICCAI-Sliver07 test set. As can be seen, our method achieves a mean score of 77.8, outperforming most of the compared methods, such as Kainm$\ddot{u}$ller (77.3), Wimmer (76.8), Linguraru (76.2), Heimann (67.6) and Kinda (64.1). In addition, the proposed method achieves the highest VOE and ASD scores.
\begin{table*}[!ht]
\begin{center}
\caption{Comparison results on the 3Dircabd database. Results are represented as
mean and standard deviation}
\label{table:Comparison on 3Dircadb}
\begin{tabular}{lccccc}
\hline
3Dircadb & VOE[\%] &RVD[\%] & ASD[mm]&RMSD[mm]&MSD[mm]\\
\hline
Chuang et al. \cite{Chung2013Regional} &12.99$\pm$5.04 &-5.66$\pm$5.59 &2.24$\pm$1.08 & - &25.74$\pm$8.85\\
Kirscher et al. \cite{Kirschner} &- &-3.62$\pm$5.50 &1.94$\pm$1.10 & 4.47$\pm$3.30 &34.60$\pm$17.70\\
Li et al. \cite{Li2015Automatic} &9.15$\pm$1.44 &-0.07$\pm$3.64 &1.55$\pm$0.39 & 3.15$\pm$0.98 &28.22$\pm$8.31\\
Erdt et al. \cite{Erdt2010Fast} &10.34$\pm$3.11 &1.55$\pm$6.49 &1.74$\pm$0.59 & 3.51$\pm$1.16 &26.83$\pm$8.87\\
\hline
3D CNNs &14.91$\pm$6.75 &-0.61$\pm$5.73 &1.86$\pm$1.86 & 5.90$\pm$3.52 &44.84$\pm$23.83\\
The Proposed&9.36$\pm$3.34 &0.97$\pm$3.26 &1.89$\pm$1.08 &4.15$\pm$3.16 &33.14$\pm$16.36\\

\hline
\end{tabular}
\end{center}
\end{table*}
\begin{figure}[!ht]
\label{fig:10}
\centering\includegraphics[width=0.4\textwidth]{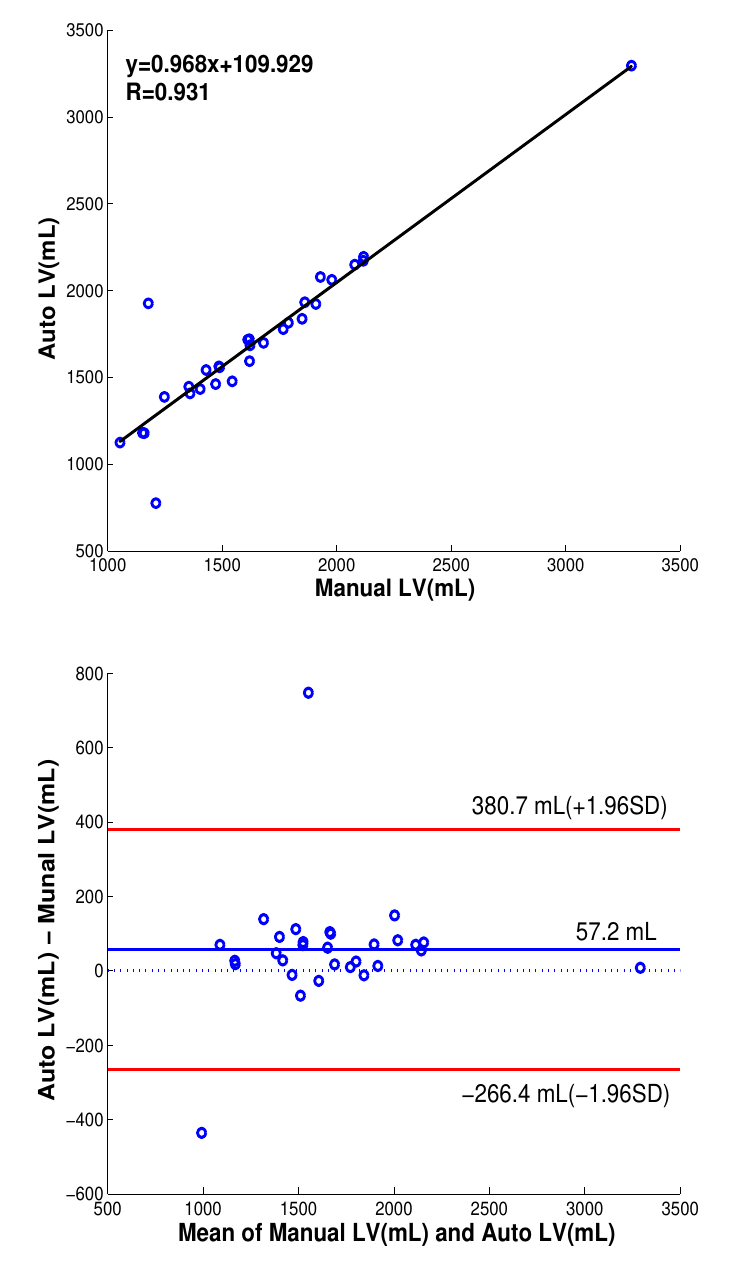}
\caption{Correlation graph (top) and Bland-Altman (bottom) for liver volume (LV)}
\end{figure}

In the second test, the results of previous methods in \cite{Chung2013Regional,Kirschner,Li2015Automatic,Erdt2010Fast}, 3D CNNs and the proposed model based on the 3Dircadb database are summarized in Table 4. Large distance between the learned liver surface and manual segmentation can be observed in terms of ASD, RMSD and MSD, as shown in the 5th row of Table 4. The proposed method achieves much better performance than Chung$'$s method except for MSD error. For most measures, the proposed method shows slightly better performance than Kirschner$'$s and Erdt$'$s. Based on shape constraints and deformable graph cut, Li$'$s method can reduce under segmentation or over segmentation of livers, and its results show slightly better performance than ours.
\par In addition, Fig. 10 illustrates the correlation graphs (top) between the segmentation and manual delineations and the Bland-Altman graphs (bottom) of the
differences, using the 10 MICCAI-Sliver07 training data and 20 3Dircadb data, for liver volume (LV). A correlation with the ground truth contours of 0.968 for LV is measured. The level of agreement between the automatic and manual results was represented by the interval of the percentage difference between mean$\pm$1.96 SD. The mean and confidence interval of the difference between the automatic and manual LV results were 57.2 mL and (-266.4 mL to 380.7 mL), respectively. The CV is 2.89. The high correlation between the automatic and manual delineations show the accuracy and clinical applicability of our method for automatic evaluation of the LV function.
\begin{figure}[!ht]
\label{fig:11}
\centering\includegraphics[width=0.4\textwidth,height=0.2\textwidth]{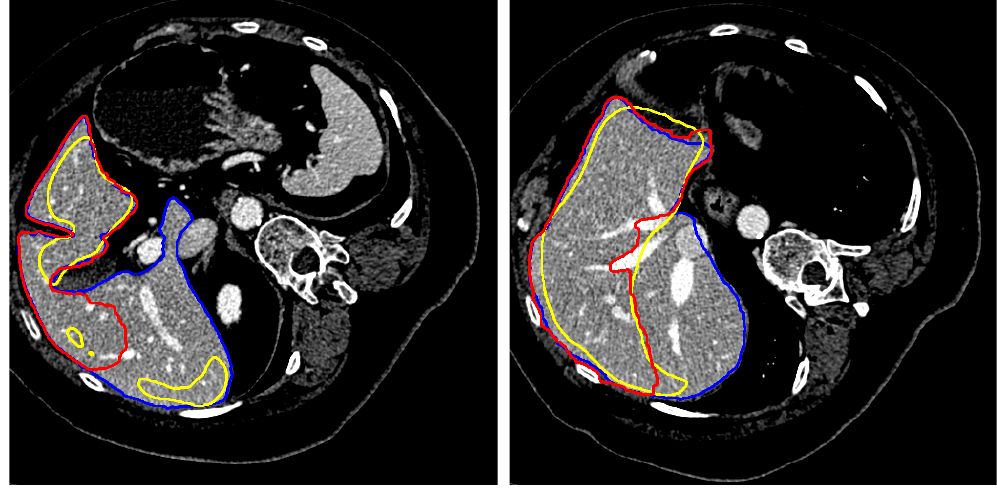}
\caption{A typical case from MICCAI-Sliver07 training set. The results of 3D CNNs and the proposed model are in yellow and red respectively. The ground truth segmentation is in blue}
\end{figure}
\begin{figure}[!ht]
\label{fig:12}
\centering\includegraphics[width=0.4\textwidth,height=0.2\textwidth]{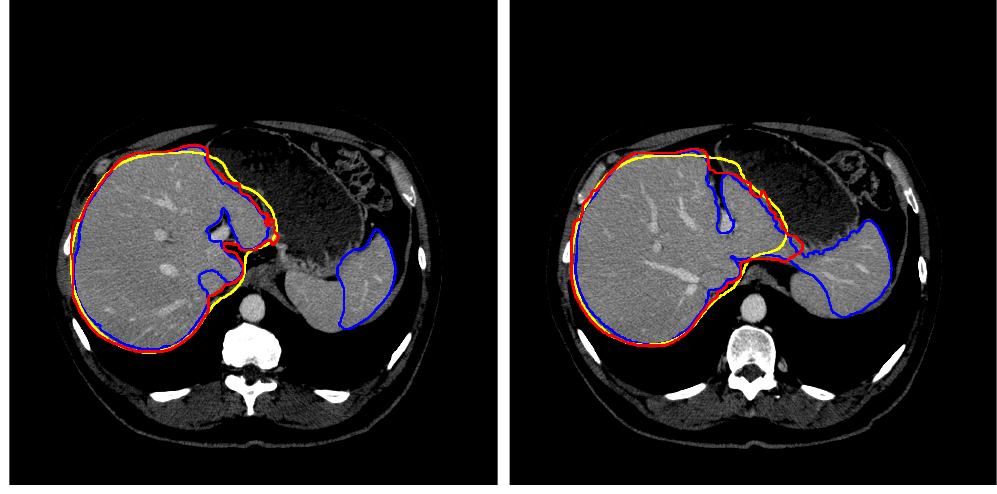}
\caption{A typical case from 3Dircadb data set. The results of 3D CNNs and the proposed method are in yellow and red respectively. The ground truth segmentation is in blue}
\end{figure}
\par Despite the overall promising results, there are also  several limitations that should be considered in
future study. Large surface distances occasionally occurs in the connection of the liver and vessels as shown in Fig.~7 and Fig. 8. In addition, several typical failure cases are shown in Fig.~11 and Fig.~12. The first case is the liver005 of MICCAI-Sliver07 training dataset, as shown in Fig.~10. This subject is laid on one side, leading to a large rotation. Our model obtained a poor segmentation since CNNs is not rotationally invariant \cite{Zeiler2014Visualizing}. In future work, this issue may be resolved by an align algorithm as a preprocessing step. The second case is from 3Dircabd database, as shown in Fig.~11. The high similarity of intensities between the left lope and its surrounding organ makes it extremely difficult to identify the left lope accurately. The under-segmentation result of this case indicates that more special characteristics of the liver¡¯s anatomical structure should be considered.
\section*{Conclusion}
\label{sec:conclusion}
In this study, we explored 3D CNNs for automatic liver segmentation in abdominal CT images. Specifically, a generative 3D CNNs model was trained for automatic liver detection. Meanwhile, a probability map of the target liver can be obtained, giving rise to an initial segmentation. The learned probability map was then integrated into the energy function of graph cut for further segmentation refinement. The main advantages of our method are that it does not require any user interaction for initialization. Thus, the proposed method can be performed by non-experts. In addition, our work is one of the early attempts of employing deep learning algorithms for 3D liver segmentation.
\par  The proposed method is evaluated on two public datasets MICCAI-Sliver07 and 3Dircabd. By comparing with state-of-the-art automatic liver segmentation methods, our method
demonstrated superior segmentation accuracy. The high correlation between our segmentation and manual references indicates that the proposed method has the clinical applicability for hepatic volume estimation. In future work, we plan to apply our method to other medical image segmentation tasks, such as kidney and spleen segmentation.


\section*{Acknowledgements}
The authors would like to thank Professor Yuan Jing for his valuable discussion and useful suggestion. This work was supported in part by National Natural Science Foundation of China (Grant Nos.: 11271323, 91330105, 11401231) and the Zhejiang Provincial Natural Science Foundation of China (Grant No.: LZ13A010002).

\section*{Compliance with ethical standards}
\textbf{Conflict of interest:}~~The authors declare that they have no conflict of
interest.\\
\textbf{Ethical standard:}~~This article does not contain any studies with human participants or animals performed by
any of the authors.\\
\textbf{Informed consent:}~~Informed consent was obtained from all individual participants included
in the study.
\bibliographystyle{spmpsci}      
\bibliography{ref}   


\end{document}